\documentclass[letterpaper, 10 pt, conference]{ieeeconf}

\IEEEoverridecommandlockouts
\usepackage{graphicx}
\usepackage{amsmath, amssymb}
\usepackage{bm}
\usepackage{siunitx}
\usepackage{booktabs}
\usepackage{threeparttable}
\usepackage{multirow}
\usepackage[hidelinks]{hyperref}
\hypersetup{
  pdftitle={Cross-Platform Control for Autonomous Surface Vehicles via Adaptive Reinforcement Learning},
  pdfauthor={Ruiheng Jiang, Thomas Bi, Raffaello D'Andrea, Aswin Ramachandran},
  pdfsubject={Autonomous Surface Vehicles; Reinforcement Learning; Adaptive Control},
  pdfkeywords={autonomous surface vehicles, ASV, reinforcement learning, adaptive control, cross-platform, teacher-student, sim-to-real, trajectory tracking}
}
\usepackage{xcolor}
\usepackage{tikz}
\usetikzlibrary{shapes.geometric, arrows.meta, positioning, fit, calc}
\usepackage{pgfplots}
\pgfplotsset{
    compat=1.18,
    every axis/.append style={
        tick label style={font=\footnotesize},
        label style={font=\footnotesize},
        legend style={font=\footnotesize},
    },
}

\title{\LARGE \bf
Cross-Platform Control for Autonomous Surface Vehicles via Adaptive Reinforcement Learning
}

\author{Ruiheng Jiang$^{1}$, Thomas Bi$^{1}$, Raffaello D'Andrea$^{1}$, and Aswin Ramachandran$^{1}$%
	\thanks{$^{1}$ Institute for Dynamic Systems and Control, ETH Zurich, Switzerland. Corresponding author: {\tt\small rjiang@ethz.ch}}%
}

\begin{document}

\definecolor{encoderblue}{RGB}{66, 133, 244}
\definecolor{policygreen}{RGB}{52, 168, 83}
\definecolor{adaptred}{RGB}{234, 67, 53}
\definecolor{envpurple}{RGB}{156, 39, 176}
\definecolor{lightgray}{RGB}{245, 245, 245}
\definecolor{darktext}{RGB}{50, 50, 50}

\maketitle
\thispagestyle{empty}
\pagestyle{empty}

\begin{abstract}
Autonomous surface vehicles vary widely in hydrodynamic and actuation characteristics, yet most controllers are designed for single-platform deployment.
We present an adaptive reinforcement learning approach for trajectory tracking that enables zero-shot cross-platform deployment using a single policy. 
Since the deployment platform's dynamics are unknown to the policy, we address cross-platform generalization with the standard partial-observability approach of conditioning on interaction history, employing a teacher--student architecture in which a learned module infers a latent representation of the platform dynamics.
The policy is trained in simulation under randomized vessel dynamics and is deployed zero-shot to two real-world platforms without any fine-tuning, despite relying on a simple analytical dynamics model rather than a high-fidelity hydrodynamic simulator.
In real-world experiments on two different platforms, the adaptive policy outperforms non-adaptive learning-based baselines by up to 58\% in position mean absolute error while approaching the tracking accuracy of a platform-specific tuned controller.

\end{abstract}

\section{Introduction}
Autonomous surface vehicles (ASVs) are increasingly deployed across a wide range of domains~\cite{manley2008usv,yan2010development}, including surveillance and maritime security, oceanographic observation~\cite{meinig2019saildrone}, environmental and water quality monitoring~\cite{dunbabin2010experimental}, disaster response and search and rescue~\cite{jorge2019survey}, hydrographic surveying and autonomous shipping~\cite{barrera2021trends}, urban waterway transportation~\cite{wang2023roboat3,homburger2025solgenia}, offshore infrastructure inspection~\cite{campos2024nautilus}, and, more recently, aquatic entertainment and robotic art~\cite{ramachandranvenkatapathy2026wayofwater}.
ASV platforms span a wide range of hydrodynamic and actuation characteristics, which increases the per-platform engineering effort required to deploy heterogeneous fleets.
A single controller that generalizes across this diversity would remove the per-platform bottleneck.

Standard approaches to ASV trajectory tracking often rely on model-based control~\cite{fossen2011handbook,liu2016unmanned,wang2018icra,kinjo2021trajectory}, which can achieve high accuracy but requires both a sufficiently accurate dynamics model obtained through system identification and platform-specific controller tuning.
Reinforcement learning (RL) offers an alternative by training in simulation~\cite{woo2019deep,wang2023deepreinforcementlearningbased,slawik2024drl}, but existing methods rely on a simulator that accurately captures the target platform's dynamics and are designed for a single, fixed vehicle rather than cross-platform deployment.
A generalist controller eliminates both requirements: it demands neither explicit knowledge of the deployment platform's dynamics nor per-platform tuning.

\begin{figure}[t]
\centering
\includegraphics[trim={0 0 51bp 0}, clip, width=\columnwidth]{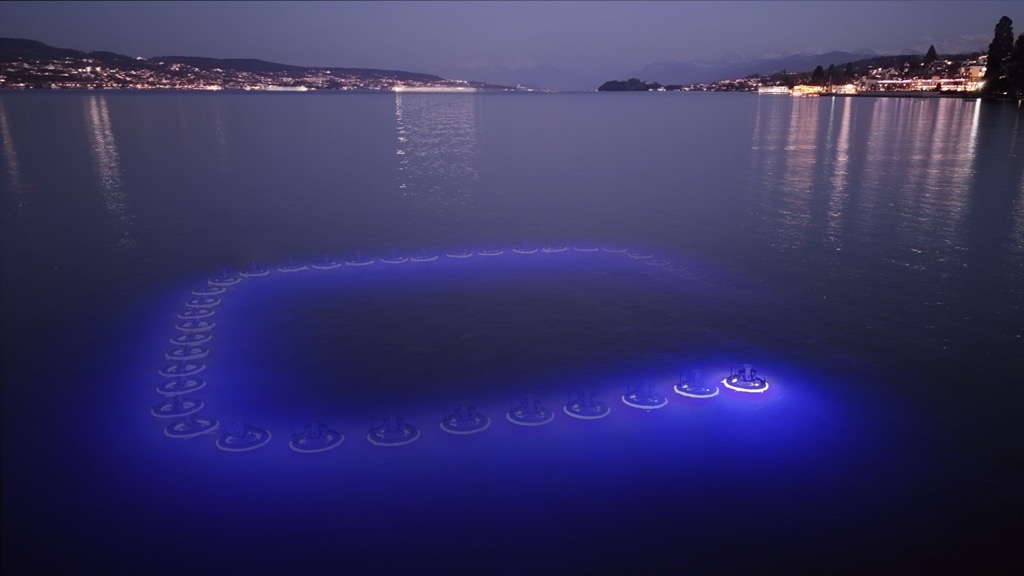}
\caption{Real-world experiment on Platform\,A, one of two ASV platforms used for real-world validation.}
\label{fig:gen1_first_page}
\end{figure}

Because the deployment platform's dynamics are unknown, cross-platform generalization is naturally a partial observability problem, for which conditioning on interaction history is a standard approach.
We propose a recipe for cross-platform ASV control:
(i)~output normalized body-frame forces to decouple the policy from absolute force magnitudes;
(ii)~train over widely randomized vessel dynamics using a lightweight analytical dynamics model for straightforward randomization and training speed;
and (iii)~equip the policy with mechanisms that infer the unknown dynamics from interaction history at deployment time.

We employ a teacher--student pipeline~\cite{Lee_2020,kumar2021rmarapidmotoradaptation}, adapted from legged locomotion, which first trains a teacher that learns a latent representation of system dynamics and then distills the teacher into a student whose adaptation module recovers the latent representation from interaction history alone.
To our knowledge, no prior work has trained a single RL policy and deployed it zero-shot across multiple marine vessels of the same class with different dynamics and actuation.

The main contribution of this work is a cross-platform adaptive control framework for ASVs: a single generalist policy that, by conditioning on interaction history, deploys zero-shot on different platforms of the same class without per-platform identification or tuning.
Training relies solely on a simple analytical dynamics model rather than high-fidelity hydrodynamic simulation. We validate this approach in real-world experiments on two different platforms of the same class (Fig.~\ref{fig:gen1_first_page}), where the adaptive policy approaches the tracking accuracy of a platform-specific tuned model predictive controller (MPC) and outperforms non-adaptive baselines by up to 58\% in position error. A supporting simulation study across five ASV platforms corroborates these findings under controlled conditions.

\section{Related Work}\label{sec:related_work}

Reinforcement learning has been applied to ASV path following~\cite{woo2019deep,slawik2024drl,zhao2021path}, collision avoidance~\cite{meyer2020colreg}, and trajectory tracking~\cite{wang2023deepreinforcementlearningbased}, with several demonstrating real-world deployment~\cite{woo2019deep,wang2023deepreinforcementlearningbased,slawik2024drl}.
However, each method is trained and validated on a single, fixed platform: the learned policy is coupled to a specific vessel, so deploying on a new platform requires re-identification of the dynamics model and retraining.

More recently, single-policy cross-platform deployment has been demonstrated across legged robots with diverse masses and morphologies~\cite{feng2023genloco,shafiee2024manyquadrupeds}; we bring this objective to the marine domain.

We treat the unknown deployment dynamics as a partial observability problem, where the prior approaches adapt to unknown dynamics by combining randomized training with online system identification or recurrent policies~\cite{yu2017uposi,peng2018sim,ni2022recurrent}.
Predicting exact dynamics parameters online is often brittle and unnecessary, while a learned latent embedding captures the information relevant for control.
A teacher--student architecture from legged locomotion~\cite{Lee_2020,kumar2021rmarapidmotoradaptation} learns such an embedding, distilling a teacher with privileged access to environment parameters into a student that infers it from interaction history alone. In contrast to purely recurrent approaches, this framework provides explicit supervision for what the adaptation module should recover, yielding a more structured and interpretable adaptation mechanism.

\section{Problem Formulation}
\label{sec:problem}

This section introduces the vessel dynamics model used for simulation and training, and formulates trajectory tracking as a partially observable Markov decision process in which the vessel's physical parameters are unknown at deployment time.
For ease of notation, vectors are expressed as $n$-tuples $(x_1, x_2, \ldots)$, with dimension and stacking clear from context.

\subsection{System Dynamics}

\begin{figure}[htbp]
  \centering
  \begin{tikzpicture}[>=Stealth]
    \draw[dashed, gray, line width=1pt] (0.75, 0.75)
      .. controls (1.30, 1.30) and (1.80, 1.90) .. (2.30, 2.50)
      .. controls (2.80, 3.10) and (3.30, 3.60) .. (3.80, 4.00)
      .. controls (4.10, 4.25) and (4.20, 4.50) .. (4.10, 4.75);

    \draw[->, line width=1pt] (0.75, 0.63) -- ++(0.88, 0) node[right] {$\bm{x}$};
    \draw[->, line width=1pt] (0.75, 0.63) -- ++(0, 0.88) node[above] {$\bm{y}$};
    \fill (0.75, 0.63) circle (1.5pt);

    \begin{scope}[shift={(3, 3.13)}, rotate=40]
      \filldraw[fill=gray!60, draw=black, line width=1pt]
        (0.625, 0) -- (0.3, 0.25) -- (-0.5, 0.2)
        -- (-0.625, 0) -- (-0.5, -0.2) -- (0.3, -0.25) -- cycle;
      \draw[fill=white, line width=0.7pt] (0.5, 0) circle (0.065);

      \draw[->, line width=1pt] (0,0) -- (1.25, 0) node[above right] {$u$};

      \draw[->, line width=1pt] (0,0) -- (0, 0.88) node[above] {$v$};

      \draw[-{Stealth[length=2.5mm, width=1.5mm]}, line width=0.8pt]
        (110:0.6) arc (110:265:0.6);
      \node at (-0.15, -0.85) {$r$};
    \end{scope}

    \draw[dashed, gray, line width=0.5pt] (3, 3.13) -- ++(1.5, 0);
    \draw[-{Stealth[length=2mm, width=1.2mm]}, line width=0.7pt]
      (3, 3.13) ++(0:0.9) arc (0:40:0.9);
    \node at ($(3, 3.13) + (15:1.15)$) {$\psi$};
  \end{tikzpicture}
  \caption{Coordinate frames and velocity components for the 3-DoF horizontal-plane model. The inertial frame $(x, y)$ is fixed, while the body-frame velocities $(u, v, r)$ denote surge, sway, and yaw rate, respectively. The heading angle $\psi$ is measured from the inertial $x$-axis to the body forward direction. The dashed line indicates the reference trajectory.}
  \label{fig:craft_trajectory}
\end{figure}
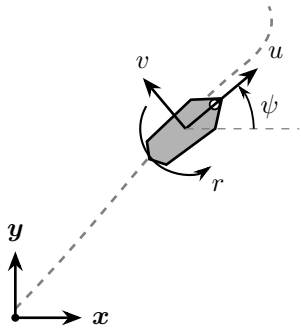

We adopt a three-degree-of-freedom (3-DoF) horizontal-plane model (Fig.~\ref{fig:craft_trajectory}) following the standard Fossen~\cite{fossen2011handbook} formulation, used for simulation, training, and inside the MPC baseline.

\subsubsection{Kinematics}
We define the system state as $\bm{s} = (\bm{\eta}, \bm{\nu})$, where the vessel pose $\bm{\eta} = (x, y, \psi)$ in the inertial frame and body-frame velocity $\bm{\nu} = (u, v, r)$ are related by
\begin{equation}
  \dot{\bm{\eta}} = \bm{J}(\psi)\,\bm{\nu},
  \qquad
  \bm{J}(\psi) =
  \begin{bmatrix}
    \cos\psi & -\sin\psi & 0 \\
    \sin\psi &  \cos\psi & 0 \\
    0 & 0 & 1
  \end{bmatrix}.
  \label{eq:kinematics}
\end{equation}

\subsubsection{Dynamics}
The equations of motion are
\begin{equation}
  \bm{M}\dot{\bm{\nu}} + \bm{N}(\bm{\nu}_r)\,\bm{\nu}_r
  = \bm{\tau} + \bm{\tau}_{\mathrm{ext}},
  \label{eq:fossen_dynamics}
\end{equation}
where $\bm{\tau} = (F_x, F_y, M_z)$ is the vector of body-frame control forces and torque, $\bm{\tau}_{\mathrm{ext}}$ represents external disturbances, and $\bm{\nu}_r = \bm{\nu} - \bm{\nu}_c$ is the velocity relative to the water current $\bm{\nu}_c$.
Each vessel is characterized by seven parameters:
\begin{equation}
\bm{\theta} = (m,\, X_{\dot{u}},\, Y_{\dot{v}},\, X_u,\, Y_v,\, N_r,\, I_{\mathrm{comb}}),
\end{equation}
where $m$ is the mass, $I_{\mathrm{comb}}$ is the combined yaw inertia, $X_{\dot{u}}, Y_{\dot{v}}$ are added-mass coefficients, and $X_u, Y_v, N_r$ are linear damping coefficients.
The inertia matrix is diagonal:
\begin{equation}
  \bm{M} = \mathrm{diag}(m - X_{\dot{u}},\; m - Y_{\dot{v}},\; I_{\mathrm{comb}}),
  \label{eq:mass_matrix}
\end{equation}
and $\bm{N}(\bm{\nu}_r) = \bm{C}(\bm{\nu}_r) + \bm{D}$ combines Coriolis--centripetal and linear damping terms:
\begin{equation}
  \bm{N}(\bm{\nu}_r) =
  \begin{bmatrix}
    X_u        & -m\,r        &  Y_{\dot{v}}\,v \\
    m\,r       &  Y_v         & -X_{\dot{u}}\,u \\
   -Y_{\dot{v}}\,v & X_{\dot{u}}\,u &  N_r
  \end{bmatrix}.
  \label{eq:hydro_matrix}
\end{equation}

\subsection{Problem Formulation}

The goal is to build a single controller that operates across a family of ASVs with diverse inertial, hydrodynamic, and actuation properties, parameterized by dynamics $\bm{\theta} \in \Theta$ that are unknown at deployment time.
We formulate trajectory tracking as a Markov decision process $(\mathcal{S}, \mathcal{A}, p, R, \gamma)$.
At each timestep, the system is in state $\bm{s}_t \in \mathcal{S}$, the agent selects action $\bm{a}_t \in \mathcal{A}$ and receives reward $R_t$, and the state transitions according to
\begin{equation}
  \bm{s}_{t+1} \sim p(\cdot \mid \bm{s}_t,\, \bm{a}_t;\, \bm{\theta}),
  \label{eq:transition}
\end{equation}
where $\bm{\theta}$ parameterizes the vessel dynamics.
While $\bm{s}_t$ is accessible via onboard sensors, $\bm{\theta}$ is not directly accessible during deployment.
Identical $(\bm{s}_t, \bm{a}_t)$ pairs can therefore produce different next states on different platforms, making the problem a partially observable Markov decision process with true Markov state $(\bm{s}_t, \bm{\theta})$. Conditioning on interaction history restores approximate Markovianity, as past state transitions and corresponding actions serve as proxies for $\bm{\theta}$.

The policy $\pi$ is optimized to maximize expected discounted return:
\begin{equation}
\pi^* = \arg\max_\pi \,\mathbb{E}_\pi\!\left[\sum_{t=0}^{\infty} \gamma^t\, R_t\right].
\end{equation}
The reference trajectory is a time series of desired states:
\begin{equation}
\bm{s}^{\mathrm{ref}}_t = (x^{\mathrm{ref}}_t,\, y^{\mathrm{ref}}_t,\, \psi^{\mathrm{ref}}_t,\, u^{\mathrm{ref}}_t,\, v^{\mathrm{ref}}_t,\, r^{\mathrm{ref}}_t).
\end{equation}

\section{Method}
\label{sec:method}

This section describes the observation and action spaces, the policy architecture, the reward, and the training procedure.

\subsection{Observation}

The base observation $\bm{o}_t \in \mathbb{R}^{6N+12}$ concatenates:
\begin{enumerate}
    \item Base features (9): body-frame position errors $(e_x^b, e_y^b)$, heading error $e_\psi$, velocity errors $(e_u, e_v, e_r)$, and velocities $(u, v, r)$.
    \item Lookahead ($6N$): $N$ future reference waypoints, each encoded as body-frame relative position $(dx^b, dy^b)$, heading difference $\Delta\psi$, reference velocities $(u_{\mathrm{ref}}, v_{\mathrm{ref}}, r_{\mathrm{ref}})$.
    \item Previous action (3): $\bar{\bm{\tau}}_{t-1}$.
\end{enumerate}

Since $\bm{o}_t$ alone does not satisfy the Markov property (Section~\ref{sec:problem}), the adaptation architecture additionally conditions on a history feature $\bm{h}_t \in \mathbb{R}^{9}$ that captures recent interaction:
\begin{equation}
\bm{h}_t = (\Delta x_t, \Delta y_t, \Delta\psi_t, u_t, v_t, r_t, \bar{\tau}_{x_{t-1}}, \bar{\tau}_{y_{t-1}}, \bar{\tau}_{\psi_{t-1}}),
\end{equation}
where $(\Delta x_t, \Delta y_t, \Delta\psi_t)$ are body-frame state difference relative to the current state, $(u_t, v_t, r_t)$ are body-frame velocities, and $(\bar{\tau}_{x_{t-1}}, \bar{\tau}_{y_{t-1}}, \bar{\tau}_{\psi_{t-1}})$ are the normalized policy outputs at the respective timestep.

\subsection{Action}

The policy outputs normalized body-frame force and torque commands $\bar{\bm{\tau}} \in [-1,1]^3$, which are scaled by platform-specific actuator limits:
\begin{equation}
\bm{\tau}_c
=
\mathrm{diag}\!\left(F_x^{\max},\, F_y^{\max},\, M_z^{\max}\right)\bar{\bm{\tau}}.
\end{equation}
This normalization decouples the policy from absolute force magnitudes. For real-world deployment, the body-frame forces and torques are mapped to individual actuators via a platform-specific allocation scheme~\cite{johansen2013allocation}.

\subsection{Teacher--Student Architecture}

\begin{figure*}[t]
\centering
\resizebox{\textwidth}{!}{%
\begin{tikzpicture}[
  box/.style={draw, rounded corners=3pt, minimum width=1.6cm, minimum height=0.7cm, align=center, font=\small},
  arrow/.style={-{Stealth[length=1.8mm]}, thick},
  titlebox/.style={draw, rounded corners=5pt, thick},
  nn/.style={rectangle, draw, rounded corners=2pt, minimum width=0.8cm, minimum height=0.5cm, font=\tiny, inner sep=2pt},
]
  \pgfdeclarelayer{background}
  \pgfsetlayers{background,main}

  \node[titlebox, minimum height=6.0cm, minimum width=3cm, fill=red!7!white] (simbox) at (-8.8,1.0) {};
  \node[anchor=north, font=\bfseries\small] at (simbox.north) {Simulation};

  \node at (-8.8,1.0) {\includegraphics[width=2.2cm]{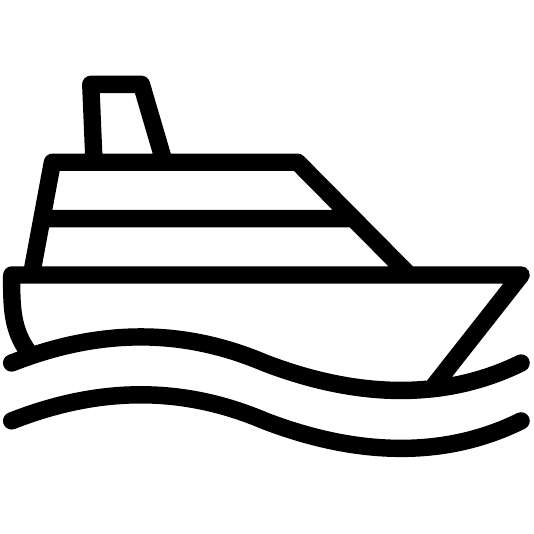}};

  \node[titlebox, minimum height=3.6cm, minimum width=12.5cm, fill=adaptred!5] (teacherbox) at (0,2.2) {};
  \node[anchor=north, font=\bfseries\small] at (teacherbox.north) {Phase 1: Teacher};

  \node[box, fill=white] (tobs) at (-4.2,3.0) {Observation $o_t$};

  \node[box, fill=adaptred!15] (priv) at (-4.2,1.4) {Privileged $\bm{e}$};

  \node[box, fill=adaptred!25, minimum width=1.8cm, minimum height=0.9cm] (enc) at (-1.6,1.4) {Encoder $\mu$};

  \node[circle, draw, thick, minimum size=0.8cm, font=\bfseries\small, fill=white] (z) at (0.4,1.4) {$\bm{z}$};

  \node[box, minimum width=2.8cm, minimum height=2.4cm, fill=envpurple!15] (tpolicy) at (3.2,2.2) {};
  \node[anchor=north, font=\small\bfseries] at (tpolicy.north) {Policy};
  \node at (3.2,1.9) {\includegraphics[height=1.3cm]{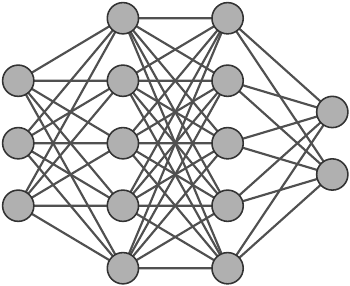}};

  \draw[arrow] (priv.east) -- (enc.west);
  \draw[arrow] (enc.east) -- (z.west);
  \draw[arrow] (tobs.east) -- ($(tpolicy.west)+(0,0.8)$);
  \draw[arrow] (z.east) -- ($(tpolicy.west)+(0,-0.8)$);

  \node[box, fill=white, minimum width=0.8cm] (taction) at (5.5,2.2) {$\bar{\bm{\tau}}_t$};
  \draw[arrow] (tpolicy.east) -- (taction.west);

  \node[titlebox, minimum height=3.6cm, minimum width=12.5cm, fill=encoderblue!5] (studentbox) at (0,-2.2) {};
  \node[anchor=north, font=\bfseries\small] at (studentbox.north) {Phase 2: Student};

  \node[box, fill=white] (sobs) at (-4.2,-1.4) {Observation $o_t$};

  \node[box, fill=encoderblue!15] (shist) at (-4.2,-3.0) {History $\bm{h}_t$};

  \node[box, fill=encoderblue!25, minimum width=1.8cm, minimum height=0.9cm] (adapt) at (-1.6,-3.0) {Adapter $\phi$};

  \node[circle, draw, thick, minimum size=0.8cm, font=\bfseries\small, fill=white] (zhat) at (0.4,-3.0) {$\hat{\bm{z}}_t$};
  \node[font=\scriptsize\itshape, below=0.05cm of zhat] {$\hat{\bm{z}}_t \!\approx\! \bm{z}$};

  \node[box, minimum width=2.8cm, minimum height=2.4cm, fill=envpurple!15, dashed] (spolicy) at (3.2,-2.2) {};
  \node[anchor=north, font=\small\bfseries] at (spolicy.north) {Policy \scriptsize\itshape(frozen)};
  \node at (3.2,-2.5) {\includegraphics[height=1.3cm]{figures/nn.pdf}};

  \draw[arrow] (shist.east) -- (adapt.west);
  \draw[arrow] (adapt.east) -- (zhat.west);
  \draw[arrow] (sobs.east) -- ($(spolicy.west)+(0,0.8)$);
  \draw[arrow] (zhat.east) -- ($(spolicy.west)+(0,-0.8)$);

  \node[box, fill=white, minimum width=0.8cm] (saction) at (5.5,-2.2) {$\bar{\bm{\tau}}_t$};
  \draw[arrow] (spolicy.east) -- (saction.west);

  \draw[arrow] (simbox.east |- teacherbox.west) -- (teacherbox.west);

  \coordinate (simstud) at (0,-1.4);
  \draw[arrow] (simbox.east |- simstud) -- (studentbox.west |- simstud);

  \draw[arrow, rounded corners=5pt] (teacherbox.east) -- ++(0.6,0) -- ++(0,2.4) -| (simbox.north);

  \draw[arrow, rounded corners=5pt] ($(studentbox.east)+(0,0.4)$) -- ++(0.6,0) -- ++(0,1.8) -- (-7.3,0);

  \node[titlebox, minimum height=3.6cm, minimum width=3cm, fill=blue!7!white] (expbox) at (9.3,-2.2) {};
  \node[anchor=north, font=\bfseries\small] at (expbox.north) {Real};
  \node at (9.3,-2.4) {\includegraphics[width=2.5cm]{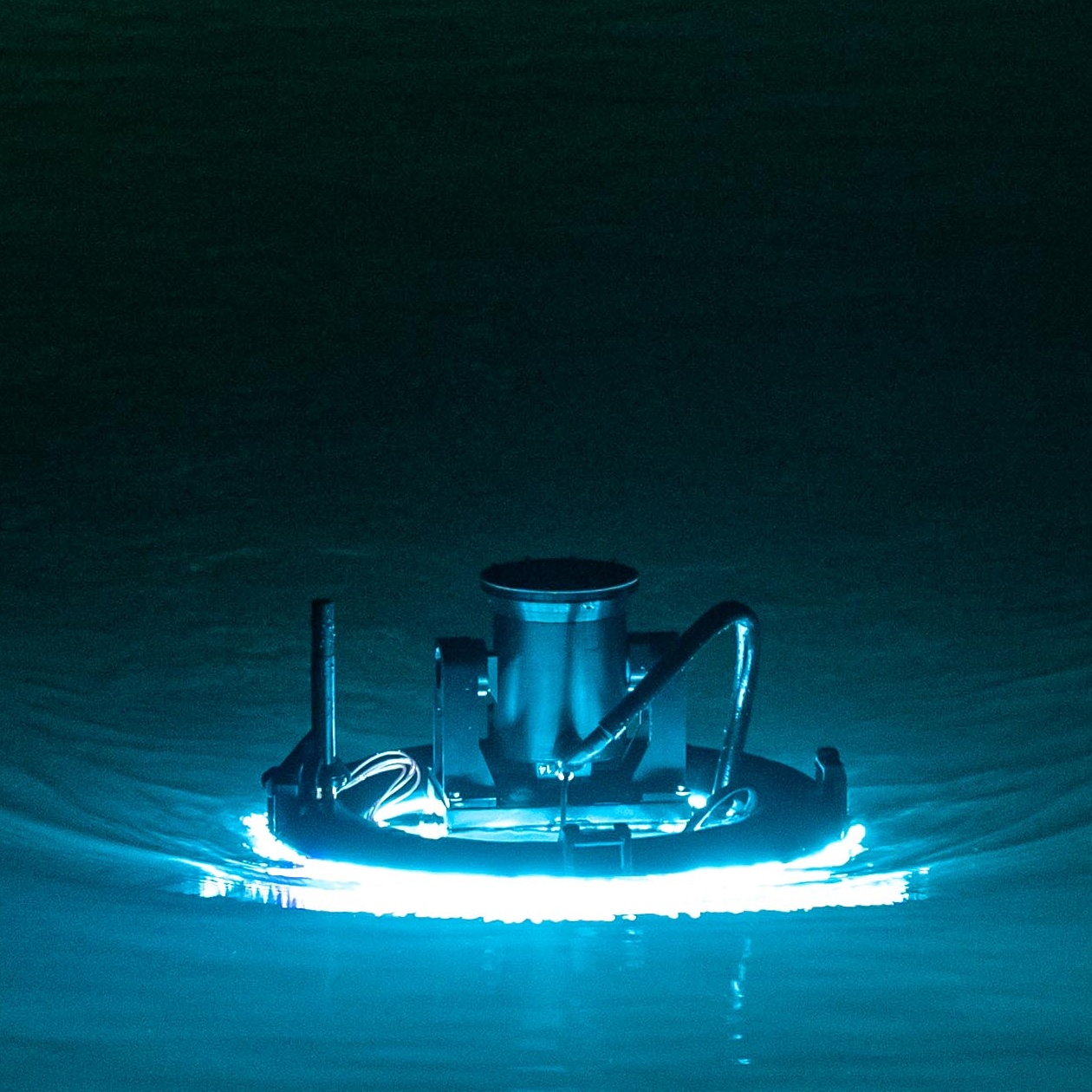}};

  \draw[arrow] ($(studentbox.east)+(0,-0.4)$) -- ($(expbox.west)+(0,-0.4)$);

  \draw[arrow, rounded corners=5pt] (expbox.south) -- ++(0,-0.6) -- (-7.0,-4.6) -- (-7.0,-3.0) -- ($(studentbox.west)+(0,-0.8)$);

  \begin{pgfonlayer}{background}
    \draw[rounded corners=8pt, fill=blue!7!white, draw=none] (-7.5,-5.0) rectangle (11.3,0.6);
    \node[anchor=north east, inner sep=8pt, font=\large\bfseries] at (11.3,0.6) {DEPLOYMENT};

    \draw[rounded corners=8pt, fill=red!7!white, draw=none] (-10.8,-2.3) rectangle (7.3,5.5);
    \node[anchor=north east, inner sep=8pt, font=\large\bfseries] at (7.3,5.5) {TRAINING};
  \end{pgfonlayer}

\end{tikzpicture}%
}
\caption{Teacher--student architecture. In Phase~1, the encoder maps the privileged dynamics vector~$\bm{e}$ to a latent~$\bm{z}$. In Phase~2, the encoder is replaced by a gated recurrent unit (GRU) adapter that estimates~$\hat{\bm{z}}$ from state-action history, while the policy weights remain frozen. Both phases train in simulation with domain randomization over the 3-DoF dynamics.}
\label{fig:teacher_student}
\end{figure*}

We adopt a two-phase approach inspired by privileged learning~\cite{Lee_2020,kumar2021rmarapidmotoradaptation} (Fig.~\ref{fig:teacher_student}). In Phase~1, a teacher policy receives a privileged vector $\bm{e} \in \mathbb{R}^{d_e}$ encoding platform dynamics. We construct $\bm{e}$ from coefficients in the 3-DoF equations of motion, written in terms of the normalized outputs $\bar{\tau}_x, \bar{\tau}_y, \bar{\tau}_\psi \in [-1,1]$:
\begin{equation}
\begin{aligned}
\dot{u} &= \alpha_u \bar{\tau}_x - \tfrac{1}{\tau_u} u + c_{rv}\, r\, v, \\
\dot{v} &= \alpha_v \bar{\tau}_y - \tfrac{1}{\tau_v} v + c_{ru}\, r\, u, \\
\dot{r} &= \alpha_r \bar{\tau}_\psi - \tfrac{1}{\tau_r} r + c_{uv}\, u\, v,
\end{aligned}
\end{equation}
where the $\alpha$ terms map normalized commands to peak accelerations, the $1/\tau$ terms represent linear hydrodynamic drag, and the $c$ terms capture Coriolis effects. This yields the nine-dimensional privileged vector:
\begin{equation}
\bm{e} = \Big(
\underbrace{\alpha_u,\, \alpha_v,\, \alpha_r}_{\text{peak accel.}},\;
\underbrace{\tfrac{1}{\tau_u},\, \tfrac{1}{\tau_v},\, \tfrac{1}{\tau_r}}_{\text{damping rates}},\;
\underbrace{c_{rv},\, c_{ru},\, c_{uv}}_{\text{Coriolis}}
\Big)
\end{equation}
with coefficients:
\begin{equation}
\begin{aligned}
\alpha_u &= \tfrac{F_x^{\max}}{m - X_{\dot{u}}}, &\quad
\alpha_v &= \tfrac{F_y^{\max}}{m - Y_{\dot{v}}}, &\quad
\alpha_r &= \tfrac{M_z^{\max}}{I_{\mathrm{comb}}}, \\[4pt]
\tfrac{1}{\tau_u} &= \tfrac{X_u}{m - X_{\dot{u}}}, &\quad
\tfrac{1}{\tau_v} &= \tfrac{Y_v}{m - Y_{\dot{v}}}, &\quad
\tfrac{1}{\tau_r} &= \tfrac{N_r}{I_{\mathrm{comb}}}, \\[4pt]
c_{rv} &= \tfrac{m - Y_{\dot{v}}}{m - X_{\dot{u}}}, &\quad
c_{ru} &= \tfrac{-(m - X_{\dot{u}})}{m - Y_{\dot{v}}}, &\quad
c_{uv} &= \tfrac{Y_{\dot{v}} - X_{\dot{u}}}{I_{\mathrm{comb}}}.
\end{aligned}
\end{equation}

An encoder $\mu: \mathbb{R}^{d_e} \rightarrow \mathbb{R}^{d_z}$ maps the privileged vector to a latent representation $\bm{z} = \mu(\bm{e})$. The actor and critic receive $[\bm{o}_t, \bm{z}]$ and output the action $\bar{\bm{\tau}}_t$ and value $V_t$.

In Phase~2, the encoder is replaced by an adapter $\phi$ that estimates $\bm{z}$ from per-step features $\bm{h}_t$ via the GRU:
\begin{equation}
\bm{g}_t = \mathrm{GRU}(\bm{h}_t,\, \bm{g}_{t-1}), \qquad
\hat{\bm{z}}_t = \mathrm{MLP}(\bm{g}_t) \in \mathbb{R}^{d_z}.
\end{equation}

In Phase~1, the policy and encoder are trained jointly with Proximal Policy Optimization (PPO)~\cite{schulman2017proximalpolicyoptimizationalgorithms} using the hyperparameters in Table~\ref{tab:training_hyperparams}.
In Phase~2, the encoder is replaced by the adapter and the actor weights are frozen. The adapter is trained by minimizing $\|\hat{\bm{z}}_t - \bm{z}\|^2$ via supervised learning using AdamW, with weight decay $10^{-5}$ and learning rate annealed linearly from $3 \times 10^{-4}$ to $3 \times 10^{-5}$.
The actor conditions on the adapter's estimate~$\hat{\bm{z}}_t$ during Phase~2 rollouts, so the adapter is trained on the state distribution induced by its own predictions rather than by the teacher's latent.
Architecture details are in Table~\ref{tab:architecture}.

\begin{table}[t]
\centering
\begin{threeparttable}
\caption{Teacher--student parameters.}
\label{tab:architecture}
\begin{tabular}{lc}
\toprule
Parameter & Value \\
\midrule
Lookahead horizon $N$ & 10 \\
Observation dim $d_o$ & $6N+12 = 72$ \\
Privileged vector dim $d_e$ & 9 \\
Latent dim $d_z$ & 9 \\
\midrule
Encoder MLP & $[128, 128]$ \\
Actor MLP & $[128, 128, 64]$ \\
Critic MLP & $[128, 128, 64]$ \\
\midrule
Adapter feature dim $d_h$ & 9 \\
GRU hidden dim & 128 \\
GRU layers & 1 \\
Projection MLP & $[128]$ \\
\bottomrule
\end{tabular}
\begin{tablenotes}[flushleft]
\footnotesize
\item The latent dimension is set equal to the privileged-vector dimension ($d_z = d_e = 9$) to avoid an information bottleneck and to remove a hyperparameter. A post-hoc analysis (Fig.~\ref{fig:scree}) shows the latent's variance concentrates in a low-dimensional subspace.
\end{tablenotes}
\end{threeparttable}
\end{table}

Both phases are trained on the 3-DoF dynamics model over a wide randomization range, so that training does not require explicit knowledge of the target platform's dynamics.
This simple setup carries several practical advantages: (i)~the lightweight model enables massive parallelization and low per-step computation cost for fast training; (ii)~randomization is straightforward over the few, physically interpretable parameters; and (iii)~no high-fidelity hydrodynamic simulator is required.

Each tracking reward term uses a bounded form:
\begin{equation}
R_i = \frac{w_i}{1 + w_i e_i^2},
\end{equation}
where $e_i$ is the tracking error and $w_i$ is the weight from Table~\ref{tab:reward_weights}. The total reward is $R_t = R_t^{\mathrm{track}} + R_t^{\mathrm{aux}}$, with
\begin{equation}
R_t^{\mathrm{track}} = R_p + R_\psi + \alpha_t (R_v + R_r),
\end{equation}
where $\alpha_t = \exp(-e_p) \exp(-0.1 |e_\psi|)$ gates velocity tracking on position and heading accuracy. The auxiliary reward is
\begin{equation}
R_t^{\mathrm{aux}} = R_{\mathrm{prox}} + R_{\mathrm{fwd}} - p_{\mathrm{sway}} - p_{\mathrm{spin}} - p_{\mathrm{act}}.
\end{equation}
The auxiliary terms comprise a proximity bonus $R_{\mathrm{prox}} = w_{\mathrm{prox}} \exp(-10\, e_p)$, a forward velocity bonus $R_{\mathrm{fwd}} = w_{\mathrm{fwd}}\, u$, quadratic penalties on lateral drift $p_{\mathrm{sway}} = w_{\mathrm{sway}}\, v^2$ and excessive yaw rate $p_{\mathrm{spin}} = w_{\mathrm{spin}} \max(0,|r|-0.3)^2$, and action regularization $p_{\mathrm{act}} = p_{\tau} + p_{\Delta\tau}$, where $p_{\tau} = w_\tau \|\bar{\bm{\tau}}_t\|^2$ and $p_{\Delta\tau} = w_{\Delta\tau} \|\bar{\bm{\tau}}_t - \bar{\bm{\tau}}_{t-1}\|^2$. Action-regularization weights are annealed from zero to their final values (Table~\ref{tab:reward_weights}) over the course of training.

\begin{table}[t]
\centering
\caption{PPO training hyperparameters.}
\label{tab:training_hyperparams}
\begin{tabular}{lc}
\toprule
Parameter & Value \\
\midrule
Total environment steps & $6\times10^7$ \\
Control timestep & 100\,ms \\
Number of environments & 1024 \\
Steps per environment & 32 \\
Mini-batches & 8 \\
Learning epochs per update & 10 \\
Discount factor $\gamma$ & 0.96 \\
Generalized advantage estimation (GAE) $\lambda$ & 0.95 \\
Clip parameter $\epsilon_{\mathrm{clip}}$ & 0.10 \\
Learning rate & $3 \times 10^{-4} \to 3 \times 10^{-5}$ \\
Entropy coefficient & 0.005 \\
Value loss coefficient & 0.5 \\
Max gradient norm & 0.7 \\
Target Kullback--Leibler (KL) divergence & 0.02 \\
Episode length & 1000 steps \\
\bottomrule
\end{tabular}
\end{table}

\paragraph{Dynamics and trajectory curriculum}
Training uses a progressive curriculum~\cite{bengio2009curriculum} that gradually increases the difficulty as the number of timesteps grows.
Dynamics parameters are sampled from a wide distribution whose ranges expand progressively with the curriculum.
Reference trajectories progress from simple to complex, starting with straight lines and large circles and progressing to tight turns. Initial state perturbations in position and orientation are introduced progressively with the curriculum.

\paragraph{Disturbance model}
Environmental disturbances comprise water currents and forces from wind and waves, both modeled as Ornstein--Uhlenbeck processes~\cite{uhlenbeck1930theory}. The inertial-frame current velocity $\bm{\nu}_c = (v_{c,x}, v_{c,y})$ evolves as:
\begin{equation}
\mathrm{d}\bm{\nu}_c = -\lambda_c \bm{\nu}_c \, \mathrm{d}t + \sigma_c \, \mathrm{d}\bm{W}_t,
\end{equation}
with decay rate $\lambda_c$ and clipped to a maximum magnitude. Body-frame forces $\bm{\tau}_{\mathrm{ext}} = (F_x^w, F_y^w, M_z^w)$ follow a similar Ornstein--Uhlenbeck process with per-platform bounds set as a fraction of the respective actuator limits, ensuring that disturbances scale appropriately across platforms.

\paragraph{Actuator modeling}
Actuator characteristics are approximated as a pure time delay $\delta$ applied to the commanded action, randomized per episode during training.

\begin{table}[t]
\centering
\caption{Reward function weights.}
\label{tab:reward_weights}
\begin{tabular}{lc}
\toprule
Term & Weight \\
\midrule
Position tracking $w_p$ & 6.0 \\
Heading tracking $w_\psi$ & 3.0 \\
Velocity tracking $w_v$ & 1.0 \\
Yaw rate tracking $w_r$ & 2.0 \\
Proximity bonus $w_{\mathrm{prox}}$ & 5.0 \\
Forward velocity $w_{\mathrm{fwd}}$ & 0.05 \\
Sway penalty $w_{\mathrm{sway}}$ & 0.05 \\
Spin penalty $w_{\mathrm{spin}}$ & 0.1 \\
Action magnitude $w_{\tau}$ & $0 \to 3$ \\
Action smoothness $w_{\Delta\tau}$ & $0 \to 10$ \\
\midrule
Failure threshold & 5\,m \\
\bottomrule
\end{tabular}
\end{table}

\section{Experimental Setup}
\label{sec:experimental_setup}

To compare controller performance, we evaluate four controllers across both real-world and simulation experiments:
(i)~a platform-specific model predictive controller (MPC) with a prediction horizon of $N=10$ steps, matching the lookahead horizon of the RL policies;
(ii)~PPO (specific), a PPO policy trained and deployed on a single platform with domain randomization~\cite{peng2018sim} around that platform's dynamics;
(iii)~PPO (general), a PPO policy trained on the full dynamics range for cross-platform deployment but without any adaptation mechanism, serving as the non-adaptive baseline; and
(iv)~Student, the teacher--student adaptation approach for cross-platform deployment.
Both PPO (specific) and PPO (general) receive only the base observation $\bm{o}_t$ without history features to isolate the contribution of adaptation from interaction history.
All policies are trained in the same vectorized simulation environment with identical reward functions and training budgets.
They are trained on a single NVIDIA RTX 4090 GPU; the full teacher--student pipeline (Phase~1 and Phase~2 combined) completes in approximately 30 minutes for $6\times10^7$ timesteps.

The simulation study includes three additional baselines:
(v)~Teacher, a Phase~1 oracle policy that receives the privileged dynamics vector $\bm{e}$ directly, establishing an upper bound on student performance;
(vi)~Recurrent, an end-to-end recurrent baseline~\cite{ni2022recurrent} in which the observation $\bm{o}_t$ is compressed by a feedforward encoder, concatenated with the history features $\bm{h}_t$, and passed into a GRU whose hidden state drives both the actor and critic heads, trained end-to-end with sequence-aware PPO without explicit supervision on the dynamics; and
(vii)~Student (avg.\ $\bm{z}$), an ablation of the student in which the adapter's output is replaced by the training-set mean latent, isolating the contribution of online adaptation.

We perform real-world experiments on two physical platforms, Platform\,A and Platform\,B (Table~\ref{tab:platforms}), which differ in dynamics and actuation.
Simulation experiments across all five platforms in Table~\ref{tab:platforms}, including Roboat\,1~\cite{wang2019roboat}, Roboat\,2~\cite{wang2020roboat2}, and Roboat\,3~\cite{wang2023roboat3}, provide supporting analysis under controlled conditions.
The Roboat dynamics parameters in Table~\ref{tab:platforms} were derived from the reported vessel specifications and adapted to fit our model parameterization.

\begin{table}[t]
\centering
\caption{Platform dynamics parameters.}
\label{tab:platforms}
\footnotesize
\setlength{\tabcolsep}{3pt}
\begin{tabular}{@{}l@{\,}r rrrrr@{}}
\toprule
& & PA & PB & R1 & R2 & R3 \\
\midrule
$m$ & \tiny{kg} & 22 & 5 & 9.2 & 80 & 1050 \\
$I_{\mathrm{comb}}$ & \tiny{kg\,m$^2$} & 1.7 & 0.7 & 1.3 & 24 & 3800 \\
\midrule
$X_{\dot{u}}$ & \tiny{kg} & $-$22 & $-$10 & $-$3.8 & $-$92 & $-$120 \\
$Y_{\dot{v}}$ & \tiny{kg} & $-$22 & $-$10 & $-$14 & $-$108 & $-$120 \\
\midrule
$X_u$ & \tiny{kg/s} & 54 & 32 & 6.0 & 38 & 126 \\
$Y_v$ & \tiny{kg/s} & 54 & 32 & 7.1 & 168 & 460 \\
$N_r$ & \tiny{kg\,m$^2$/s} & 1.7 & 1.1 & 0.8 & 16 & 160 \\
\midrule
$F_x^{\max}$ & \tiny{N} & 40 & 50 & 40 & 100 & 1500 \\
$F_y^{\max}$ & \tiny{N} & 40 & 50 & 40 & 100 & 500 \\
$M_z^{\max}$ & \tiny{N\,m} & 20 & 20 & 27 & 150 & 1500 \\
\bottomrule
\end{tabular}
\\[2pt]
\scriptsize PA = Platform\,A, PB = Platform\,B, R1--R3 = Roboat\,1--3.
\end{table}

Table~\ref{tab:priv_vector_mapping} reports the ranges of normalized dynamics components encountered by the cross-platform policies during training, including acceleration capabilities ($\alpha_u, \alpha_v, \alpha_r$), inverse time constants ($1/\tau_u, 1/\tau_v, 1/\tau_r$), and Coriolis coupling coefficients ($c_{rv}, c_{ru}, c_{uv}$).

\begin{table}[t]
\centering
\caption{Normalized dynamics component ranges during training.}
\label{tab:priv_vector_mapping}
\footnotesize
\setlength{\tabcolsep}{4pt}
\begin{tabular}{@{}lc@{}}
\toprule
Component & Range \\
\midrule
$\alpha_u = F_x^{\max}/(m - X_{\dot{u}})$ & $[0.15, 12]$ m/s$^2$ \\
$\alpha_v = F_y^{\max}/(m - Y_{\dot{v}})$ & $[0.11, 11]$ m/s$^2$ \\
$\alpha_r = M_z^{\max}/I_{\mathrm{comb}}$ & $[0.10, 86]$ rad/s$^2$ \\
\midrule
$1/\tau_u = X_u/(m - X_{\dot{u}})$ & $[0.007, 34]$ s$^{-1}$ \\
$1/\tau_v = Y_v/(m - Y_{\dot{v}})$ & $[0.019, 34]$ s$^{-1}$ \\
$1/\tau_r = N_r/I_{\mathrm{comb}}$ & $[0.003, 25]$ s$^{-1}$ \\
\midrule
$c_{rv} = (m - Y_{\dot{v}})/(m - X_{\dot{u}})$ & $[0.06, 29]$ \\
$c_{ru} = -(m - X_{\dot{u}})/(m - Y_{\dot{v}})$ & $[-16, -0.03]$ \\
$c_{uv} = (Y_{\dot{v}} - X_{\dot{u}})/I_{\mathrm{comb}}$ & $[-218, 218]$ m$^{-2}$ \\
\bottomrule
\end{tabular}
\end{table} 

\section{Results and Analysis}

\subsection{Real-World Experiments}

\begin{table}[t]
\centering
\caption{Real-world trajectory-tracking results on Platform\,A and Platform\,B. Values are mean absolute error\,$\pm$\,standard deviation.}
\label{tab:realworld_results}
\footnotesize
\setlength{\tabcolsep}{3pt}
\begin{tabular}{llcc}
\toprule
Platform & Method & Pos.\ [cm] & Head.\ [$^\circ$] \\
\midrule
\multirow{4}{*}{Platform\,A}
 & MPC              & $4.6 \pm 2.5$  & $1.9 \pm 2.1$   \\
 & Student          & $4.0 \pm 2.1$  & $2.5 \pm 1.9$   \\
 & PPO (specific)   & $5.5 \pm 3.2$  & $4.0 \pm 3.0$   \\
 & PPO (general)    & $5.6 \pm 3.0$  & $2.6 \pm 1.9$   \\
\midrule
\multirow{4}{*}{Platform\,B}
 & MPC              & $4.1 \pm 2.6$ & $2.9 \pm 3.9$ \\
 & Student          & $4.0 \pm 2.4$ & $2.0 \pm 1.5$ \\
 & PPO (specific)   & $5.3 \pm 2.8$ & $2.1 \pm 2.6$ \\
 & PPO (general)    & $9.5 \pm 3.4$ & $2.2 \pm 1.6$ \\
\bottomrule
\end{tabular}
\end{table}

All RL policies were trained with wind and waves ($10\%$ of control authority), currents up to $0.2$\,m/s, and actuator delay randomized in $[0, 200]$\,ms. They were then deployed zero-shot on Platform\,A and Platform\,B without any platform-specific retraining.
The validation trajectory is a rectangle with rounded corners traversed at $0.6$\,m/s, covering both straight and curved segments representative of common ASV operational patterns.
Both platforms are equipped with a real-time kinematic (RTK) GPS running at 7\,Hz and an inertial measurement unit (IMU) at 100\,Hz; the control loop executes at 10\,Hz.
The tracking accuracy is measured as the difference between the reference and the estimated state using measurements from those onboard sensors. Results are reported in Table~\ref{tab:realworld_results}.

\subsubsection{Platform\texorpdfstring{\,}{ }A}
The student achieves the lowest position error ($4.0$\,cm), followed by MPC ($4.6$\,cm), while MPC leads in heading ($1.9^\circ$ vs.\ $2.5^\circ$).
Both outperform PPO (specific) in heading ($4.0^\circ$) and position ($5.5$\,cm).
Relative to PPO (general), the student reduces position error by $29\%$ ($4.0$\,cm vs.\ $5.6$\,cm) with comparable heading performance ($2.5^\circ$ vs.\ $2.6^\circ$).

\subsubsection{Platform\texorpdfstring{\,}{ }B}
On Platform\,B, the student matches MPC in position ($4.0$\,cm vs.\ $4.1$\,cm) and shows lower heading error ($2.0^\circ$ vs.\ $2.9^\circ$).
Compared to PPO (general), the student reduces position error by $58\%$ ($4.0$\,cm vs.\ $9.5$\,cm) while heading remains comparable ($2.0^\circ$ vs.\ $2.2^\circ$).

Across both platforms, the student consistently reduces position error relative to PPO (general), while heading errors remain comparable among the general policies.
The student achieves equal position accuracy ($4.0$\,cm) on both platforms despite their different dynamics, with a larger improvement on Platform\,B compared with PPO (general).
We attribute this to Platform\,B's higher thrust-to-weight ratio: on such a platform, suboptimal commands produce proportionally larger tracking deviations, amplifying the benefit of adaptation.
Notably, PPO (specific) does not outperform the adaptive student: without adaptation and with a narrower training distribution, it cannot compensate for the sim-to-real gap, leading to weaker performance.

\subsection{Simulation Experiments}

\begin{table}[t]
\centering
\caption{Simulation tracking under ideal conditions across all five platforms. Values are mean absolute error\,$\pm$\,standard deviation across 10 seeds. $^\star$Validated on the designated platform only.}
\label{tab:controller_comparison}
\footnotesize
\setlength{\tabcolsep}{4pt}
\begin{tabular}{lcc}
\toprule
Method & Pos.\ [cm] $\downarrow$ & Head.\ [$^\circ$] \\
\midrule
MPC$^\star$               & $0.33 \pm 0.00$ & $0.07 \pm 0.00$ \\
PPO (specific)$^\star$    & $0.65 \pm 0.04$ & $0.97 \pm 0.13$ \\
Student                   & $0.84 \pm 0.11$ & $1.81 \pm 0.50$ \\
Teacher                   & $0.85 \pm 0.13$ & $1.79 \pm 0.50$ \\
Recurrent                 & $0.96 \pm 0.31$ & $2.34 \pm 0.42$ \\
PPO (general)             & $1.49 \pm 0.10$ & $3.98 \pm 0.28$ \\
Student avg.\ $\bm{z}$ & $2.18 \pm 0.16$ & $4.06 \pm 0.37$ \\
\bottomrule
\end{tabular}
\end{table}

The simulation study extends the real-world findings to five vessels (Table~\ref{tab:platforms}) under controlled conditions.
All policies are trained and evaluated under ideal conditions (no delay, no disturbances) in Table~\ref{tab:controller_comparison}, characterizing nominal tracking ability, while robustness is assessed in the real-world experiments under external disturbances.
MPC achieves the lowest error ($0.33$\,cm), as expected when the controller operates on a perfect model of the simulated dynamics.
PPO (specific) ($0.65$\,cm) also performs well, because it is trained and validated on the same conditions without domain mismatch.
The student reduces position error by $44\%$ ($0.84$\,cm vs.\ $1.49$\,cm) and heading error by $55\%$ ($1.81^\circ$ vs.\ $3.98^\circ$) relative to PPO (general).
The larger heading improvement compared with the real-world results reflects the wider spread of yaw control authorities across the simulated platforms, where adaptation has a stronger effect.

The recurrent policy achieves $0.96$\,cm and $2.34^\circ$, improving over PPO (general) but remaining above the student ($0.84$\,cm, $1.81^\circ$).
Beyond tracking performance, the teacher--student framework yields an interpretable latent that directly encodes platform dynamics. Fixing the latent to its training-set mean (Student avg.\ $\bm{z}$) more than doubles tracking error ($2.18$\,cm, $4.06^\circ$), confirming the contribution of the adaptation module.

\begin{figure}[t]
    \centering
    \includegraphics[width=0.85\linewidth]{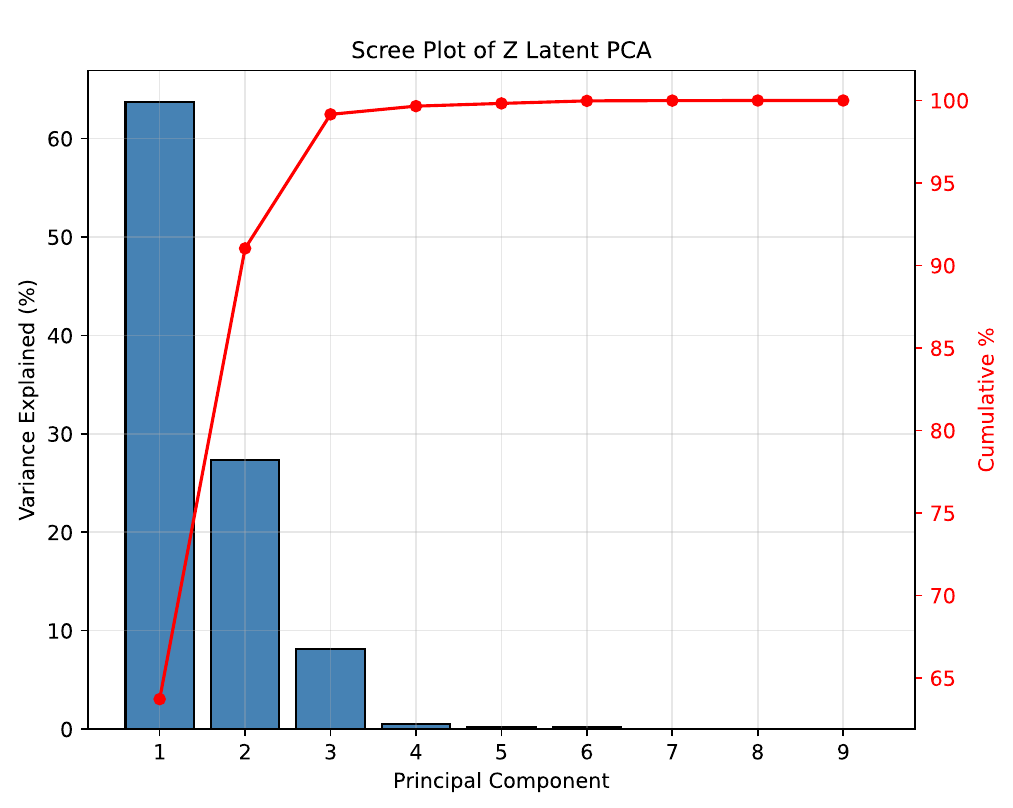}
    \caption{Scree plot of the learned latent $\bm{z}$ over the dynamics. The first two principal components capture $91\%$ of the variance, indicating that the nine-dimensional latent occupies a low-dimensional subspace and that matching $d_z$ to $d_e$ is not a binding constraint.}
    \label{fig:scree}
\end{figure}

Fig.~\ref{fig:scree} shows a principal component analysis of the learned latent across dynamics: the first two components explain $91\%$ of the variance (PC1 $63.7\%$, PC2 $27.3\%$), so the nine-dimensional latent effectively occupies a low-dimensional subspace.
Note that the effective dimensionality is also task-dependent: trained end-to-end with the policy, the encoder represents the dynamics variation that affects tracking on the training trajectories.
A controlled single-axis illustration of the adaptation mechanism for a separate variant policy is provided in Appendix~\ref{app:latent_sweep}.

\section{Conclusion}
We presented a single adaptive trajectory-tracking policy that is deployed zero-shot across platforms by conditioning on interaction history.
Deployment on a new platform requires neither explicit knowledge of the platform's dynamics nor any retraining.
In real-world experiments on two different platforms, the adaptive policy reduces position error by up to 58\% relative to the non-adaptive baseline, approaching the tracking accuracy of a platform-specific tuned MPC.
Notably, the entire training pipeline uses only a lightweight 3-DoF analytical model and completes in about 30 minutes on a single consumer GPU.

\textit{Limitations.} Our experiments cover a low-speed displacement regime, where the 3-DoF linear-damping model captures the dominant dynamics. The adapter offers limited ability to accommodate velocity-dependent effects such as quadratic drag, which at a given speed appear as an effective damping within the randomized range (Table~\ref{tab:priv_vector_mapping}) that it re-estimates online. Regimes the 3-DoF model omits, such as planing or strong heave, roll, and pitch coupling, remain future work.

Beyond broadening this operating envelope, the approach could be augmented with online or continual adaptation, allowing the policy to progressively specialize into a platform-specific expert controller, improving performance as more operational data from the deployed system becomes available.
A second direction is tighter integration with model-based control: a learned module could condition the parameters and objectives online, combining the constraint-handling and interpretability of model-based methods with the flexibility of learning-based adaptation.

\appendix
\section{Latent-Space Adaptation under a Controlled Dynamics Sweep}
\label{app:latent_sweep}
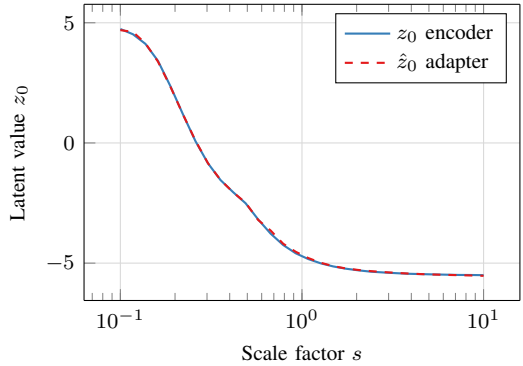
\begin{figure}[h]
    \centering
    \begin{tikzpicture}
    \begin{semilogxaxis}[
        width=0.85\linewidth,
        height=5.5cm,
        axis y line*=left,
        xlabel={Scale factor $s$},
        ylabel={Latent value $z_0$},
        grid=major,
        grid style={gray!30},
        legend style={
            at={(0.97,0.97)},
            anchor=north east,
            cells={anchor=west},
        },
        every axis plot/.append style={line width=1pt},
    ]
        \addplot[color={rgb,255:red,55;green,126;blue,184}, thick]
            table[x=s, y=z_true_0, col sep=comma] {figures/z_vs_scale.csv};
        \addplot[color={rgb,255:red,228;green,26;blue,28}, thick, dashed]
            table[x=s, y=z_hat_0, col sep=comma] {figures/z_vs_scale.csv};
        \legend{$z_0$ encoder, $\hat{z}_0$ adapter}
    \end{semilogxaxis}
    \begin{semilogxaxis}[
        width=0.85\linewidth,
        height=5.5cm,
        axis y line*=right,
        axis x line=none,
        ylabel={\phantom{Latent value $z_0$}},
        y label style={opacity=0},
        xmin=0.1, xmax=10,
        ymin=-6, ymax=2,
        ytick={-6,-4,-2,0,2},
        yticklabel style={text opacity=0},
        y tick style={draw=none},
    ]
    \end{semilogxaxis}
    \end{tikzpicture}
    \caption{Teacher latent component $z_0$ as a function of the uniform dynamics-scaling factor $s$, for the controlled one-dimensional sweep variant.}
    \label{fig:latent_vs_scale}
\end{figure}

\begin{figure}[h]
    \centering
    \begin{tikzpicture}
      \begin{axis}[
          name=mainplot,
          width=0.85\linewidth,
          height=5.5cm,
          xlabel={Time (s)},
          ylabel={Scale factor $s$},
          axis y line*=left,
          ymin=0, ymax=1.15,
          ytick={0, 0.3, 0.5, 1.0},
          xmin=5, xmax=15,
          grid=major,
          grid style={gray!30},
          every axis plot/.append style={line width=1pt},
          legend style={
              at={(0.03,0.7)},
              anchor=west,
              cells={anchor=west},
          },
      ]
        \addplot[black, very thick, const plot]
            table[x=t, y=s, col sep=comma] {figures/midrun_adaptation.csv};
        \addlegendentry{$s$}
      \end{axis}
      \begin{axis}[
          width=0.85\linewidth,
          height=5.5cm,
          axis y line*=right,
          axis x line=none,
          ylabel={Latent value $z_0$},
          ymin=-6, ymax=2,
          ytick={-6,-4,-2,0,2},
          xmin=5, xmax=15,
          grid=none,
          every axis plot/.append style={line width=1pt},
          legend style={
              at={(0.97,0.97)},
              anchor=north east,
              cells={anchor=west},
          },
      ]
        \addplot[color={rgb,255:red,55;green,126;blue,184}, thick, const plot]
            table[x=t, y=z_true_0, col sep=comma] {figures/midrun_adaptation.csv};
        \addlegendentry{$z_0$ encoder}
        \addplot[color={rgb,255:red,228;green,26;blue,28}, thick, dashed]
            table[x=t, y=z_hat_0, col sep=comma] {figures/midrun_adaptation.csv};
        \addlegendentry{$\hat{z}_0$ adapter}
      \end{axis}
    \end{tikzpicture}
    \caption{Online adaptation of the student's inferred latent $\hat{z}_0$ following an abrupt mid-trajectory change in dynamics at $t=10\,\mathrm{s}$, for the controlled one-dimensional sweep variant.}
    \label{fig:midrun_adaptation}
\end{figure}
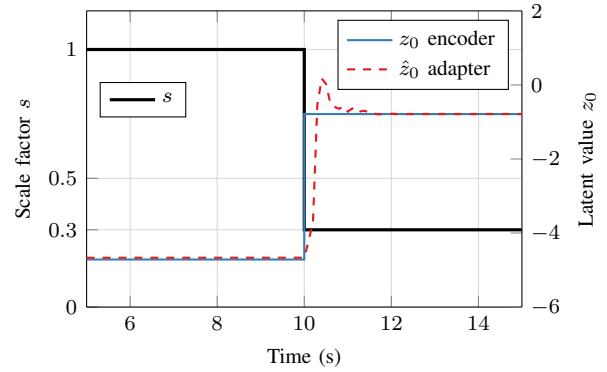

To illustrate the adaptation mechanism, we train a separate variant policy on a controlled one-dimensional dynamics sweep, in which a single factor $s$ uniformly scales all inertial and damping parameters while actuation limits remain fixed. Because the variation is one-dimensional, a single latent component $z_0$ suffices to capture it. Fig.~\ref{fig:latent_vs_scale} shows that the teacher's latent $z_0$ varies smoothly and monotonically with $s$, so the encoder learns a continuous, physically meaningful embedding of the dynamics. Reducing $s$ abruptly from $1.0$ to $0.3$ at $t=10\,\mathrm{s}$, Fig.~\ref{fig:midrun_adaptation} shows the student's inferred $\hat{z}_0$, recovered online from interaction history alone, converging within approximately $1.0\,\mathrm{s}$, illustrating the adaptation mechanism in isolation from other factors.

\section*{Acknowledgments}
The authors thank Sebastian Burmester, Jan Kamm, and Noa Sendlhofer for their help with experimental testing and for valuable discussions.

\bibliographystyle{IEEEtran}
\bibliography{bib/references}

\end{document}